\documentclass[runningheads]{llncs}
\usepackage{graphicx}
\usepackage[misc]{ifsym}
\usepackage{amsmath}
\usepackage{amsfonts}
\usepackage{graphicx}
\usepackage{multirow}
\usepackage{booktabs}
\usepackage{threeparttable}
\usepackage{hyperref}
\usepackage{subfigure}

\makeatletter
\newcommand{\printfnsymbol}[1]{%
	\textsuperscript{\@fnsymbol{#1}}%
}
\makeatother

\begin{document}
\title{TransEdge: Translating Relation-contextualized Embeddings for Knowledge Graphs}
\titlerunning{Translating Relation-contextualized Embeddings for Knowledge Graphs}
%
\author{
	Zequn Sun\inst{1}\and
	Jiacheng Huang\inst{1}\and
	Wei Hu\inst{1}\textsuperscript{(\Letter)}\and
	Muhao Chen\inst{2}\and \\
	Lingbing Guo\inst{1}\and
	Yuzhong Qu\inst{1}
}
\authorrunning{Z. Sun et al.}
%
\institute{
	State Key Laboratory for Novel Software Technology, \\ Nanjing University, Nanjing, Jiangsu, China \\
	\email{\{zqsun.nju,jchuang.nju,lbguo.nju\}@gmail.com, \{whu,yzqu\}@nju.edu.cn} \and
	Department of Computer Science, \\ University of California, Los Angeles, CA, USA \\
	\email{muhaochen@ucla.edu}
}
\maketitle              
\begin{abstract}
Learning knowledge graph (KG) embeddings has received increasing attention in recent years. Most embedding models in literature interpret relations as linear or bilinear mapping functions to operate on entity embeddings. However, we find that such relation-level modeling cannot capture the diverse relational structures of KGs well. In this paper, we propose a novel edge-centric embedding model TransEdge, which contextualizes relation representations in terms of specific head-tail entity pairs. We refer to such contextualized representations of a relation as edge embeddings and interpret them as translations between entity embeddings. TransEdge achieves promising performance on different prediction tasks. Our experiments on benchmark datasets indicate that it obtains the state-of-the-art results on embedding-based entity alignment. We also show that TransEdge is complementary with conventional entity alignment methods. Moreover, it shows very competitive performance on link prediction. 

\end{abstract}
\section{Introduction}
\begin{figure}[!t]
	\centering
	\includegraphics[width=0.99\textwidth]{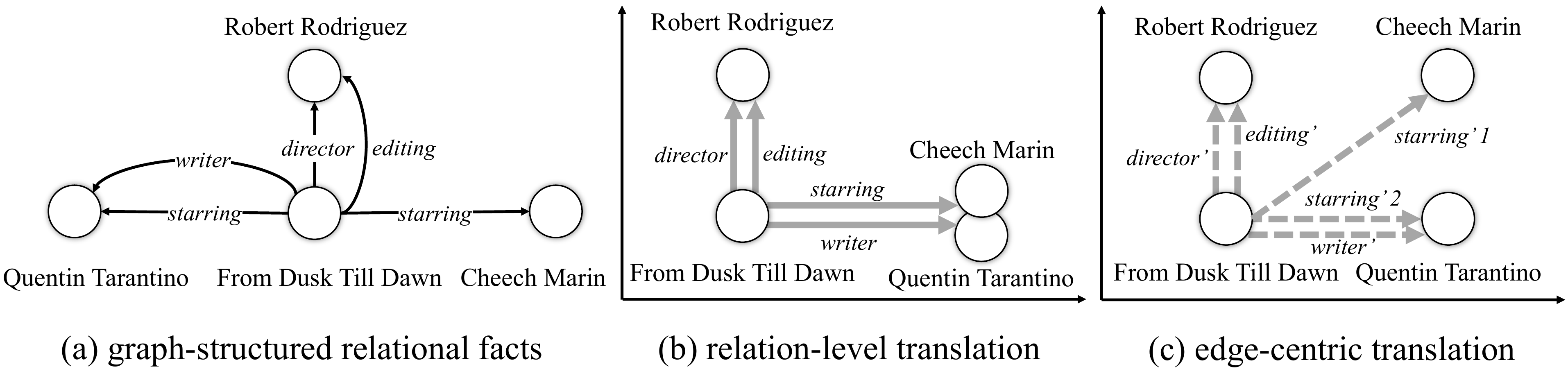}
	\caption{(a) A snapshot of the relational facts of ``From Dusk Till Dawn'' in DBpedia. Circles represent entities and directed edges have labels. (b) Illustration of the relation-level translation between entity embeddings, where circles represent entity embeddings, and bold gray arrows denote the translation vectors of relations. (c) Illustration of the the proposed edge-centric translation, where the dotted arrows denote the contextualized representations, i.e., edge embeddings. For example, \textit{starring'} 1 and \emph{starring'} 2 are two contextualized representations of the relation \textit{starring}.}
	\label{fig:example}
\end{figure}

A knowledge graph (KG) is a multi-relational graph, whose nodes correspond to entities and directed edges indicate the specific relations between entities. For example, Fig.~\ref{fig:example}(a) shows a snapshot of the graph-structured relational triples in DBpedia. In KGs, each labeled edge is usually represented by a relational triple in the form of $(\mathtt{head}, \mathtt{relation}, \mathtt{tail})$\footnote{In the following, $(\mathtt{head}, \mathtt{relation}, \mathtt{tail})$ is abbreviated as $(h,r,t)$.}, meaning that the two entities $\mathtt{head}$ and $\mathtt{tail}$ hold a specific $\mathtt{relation}$. So, a typical KG can be defined as a triple $\mathcal{K}=(\mathcal{E},\mathcal{R},\mathcal{T})$, where $\mathcal{E}$ is the set of entities (i.e., nodes), $\mathcal{R}$ is the set of relations (i.e., edge labels),~and~$\mathcal{T}=\mathcal{E}\times\mathcal{R}\times\mathcal{E}$ denotes the set of relational triples (i.e., labeled edges). Each entity or relation is usually denoted by a URI. For example, the URI of New Zealand in DBpedia is $\mathtt{dbr:New\_Zealand}$\footnote{\url{http://dbpedia.org/resource/New_Zealand}}. However, such discrete and symbolic representations of KGs fall short of supporting the efficient knowledge inference~\cite{KGE_survey}. Thus, learning continuous and low-dimensional embedding representations for KGs has drawn much attention in recent years and facilitated many KG-related tasks, such as link prediction~\cite{TransE,ConvE,SimplE,TransR,RotatE,ComplEx,TransH,DistMult,CrossE}, entity alignment~\cite{KDCoE,MTransE,EA_Table,JAPE,BootEA,GCN-Align} and entity classification tasks~\cite{Global_embed,TypeConstrain,RDF2Vec}

KG embedding seeks to encode the entities and relations into vector spaces, and capture semantics by the geometric properties of embeddings. To model the relational structures of KGs, most embedding models in literature interpret relations as linear or bilinear mapping functions operating on entity embeddings, such as the relation translation in TransE~\cite{TransE}, the relation matrix factorization in DistMult~\cite{DistMult}, and the relation rotation in RotatE~\cite{RotatE}. We refer to this kind of models as relation-level embedding. However, such relation-level models represent each relation with one embedding representation for all related head-tail entity pairs, which cannot well reflect the complex relational structures of KGs. As shown in Fig.~\ref{fig:example}(a), different entity pairs may share the same relation while one entity pair may hold different relations. The relation-level embedding cannot distinguish the different contexts of relations, which would lead to indistinguishable embeddings and incorrect relation inference.

Specifically, we take the translational KG embedding model TransE~\cite{TransE} as an example to explicate the aforementioned issue. TransE interprets relations as translation vectors between entity embeddings. For example, given a relational triple $(h,r,t)$, TransE expects $\mathbf{h}+\mathbf{r}\approx\mathbf{t}$ to hold, where the boldfaced letters denote the embeddings of entities and relations. The relation embedding $\mathbf{r}$ serves as a translation vector from $\mathbf{h}$ to $\mathbf{t}$. However, such relation translation encounters issues when facing more complex relations. For example, considering the relational triples in Fig.~\ref{fig:example}(a): (From Dusk Till Dawn, \textit{starring}, Quentin Tarantino) and (From Dusk Till Dawn, \textit{starring}, Cheech Marin), translational KG embeddings would have $ \textbf{Quentin Tarantino} \approx \textbf{Cheech Marin}$, as shown in Fig.~\ref{fig:example}(b). In other words, the different entities getting involved in the same relation would be embedded very closely by the same relation translation. Such indistinguishable entity embeddings go against accurate embedding-based entity alignment. Quentin Tarantino and Cheech Marin would be mistaken for an aligned entity pair due to the high similarity of their embeddings. Besides, the similar relation embeddings, such as $ \textbf{starring} \approx \textbf{writer}$, would lead to the incorrect link prediction such as (From Dusk Till Dawn, \textit{writer}, Cheech Marin). This problem has been noticed in the link prediction scenario~\cite{TransD,TransR,TransH}. Towards link prediction that predicts the missing entities for relational triples, they propose to distinguish entity embeddings with relation-specific projections. However, such projections divest KG embeddings of relational structures by injecting ambiguity into entity embeddings. 

In this paper, we introduce an edge-centric translational embedding model TransEdge, which differentiates the representations of a relation between different entity-specific contexts. This idea is motivated by the graph structures of KGs. Let us see Fig.~\ref{fig:example}(a). One head-tail entity pair can hold different relations, i.e, one edge can have different labels. Also, different edges can have the same label, indicating that there are multiple head-tail entity pairs having the same relation. Thus, it is intuitive that entities should have explicit embeddings while relations should have different contextualized representations when translating between different head-tail entity pairs. Thus, we propose to contextualize relations as different edge embeddings. The context of a relation is specified by its head and tail entities. We study two different methods, i.e., context compression and context projection, for computing edge embeddings given the edge direction (head and tail entity embeddings) and edge label (relation embeddings). To capture the KG structures, we follow the idea of translational KG embeddings and build translations between entity embeddings with edge embeddings. This modeling is simple but has appropriate geometric interpretations as shown in Fig.~\ref{fig:example}(c). Our main contributions are listed as follows:

\begin{itemize}
	\item[(1)] We propose a novel KG embedding model TransEdge. Different from existing models that learn one simple embedding per relation, TransEdge learns KG embeddings by contextualizing relation representations in terms of the specific head-tail entity pairs. We refer to such contextualized representations of a relation as edge embeddings and build edge translations between entity embeddings to capture the relational structures of KGs. TransEdge provides a novel perspective for KG embedding. (Section~\ref{sec:model})
	\item[(2)] We evaluate TransEdge on two tasks: entity alignment between two KGs and link prediction in a single KG. Experimental results on five datasets show that TransEdge obtains the state-of-the-art results on entity alignment. It also achieves very competitive performance (even the best Hits@1) on link prediction with low computational complexity. These experiments verify the good generalization of TransEdge. To the best of our knowledge, TransEdge is the first KG embedding model that achieves the best Hits@1 performance on both embedding-based entity alignment and link prediction. (Section~\ref{sec:exp})
\end{itemize}

\section{Related Work}
\label{sec:rw}
In recent years, various KG embedding models have been proposed. The most popular task to evaluate KG embeddings is link prediction. Besides, embedding-based entity alignment also draws much attention recently. In this section, we discuss these two lines of related work.

\subsection{KG Embeddings Evaluated by Link Prediction}
We divide existing KG embedding models evaluated by link prediction into three categories, i.e., \emph{translational}, \emph{bilinear} and \emph{neural} models. TransE~\cite{TransE} introduces the translational KG embeddings. It interprets relations as translation vectors operating on entity embeddings. Given a relational triple $(h,r,t)$, TransE defines the following energy function to measure the error of relation translation: $f_{\text{\scriptsize TransE}}(h,r,t)=||\mathbf{h} + \mathbf{r} -\mathbf{t}||$, where $||\cdot||$ denotes either the $L_1$ or $L_2$ vector norm. To resolve the issues of TransE on modeling complex relations, some improved translational models have been put forward, including TransH~\cite{TransH}, TransR~\cite{TransR} and TransD~\cite{TransD}. Their key idea is to let entities have relation-specific embeddings by transformations operating on entity embeddings, such as the hyperplane projection in TransH and the space projection in TransR and TransD. We argue that such transformations introduce ambiguity to entity embeddings as they separate the original entity embedding into many dispersive relation-specific representations. For example, for each relation $r$, entity $h$ would hold a representation $\mathbf{h}_r$. These dispersive representations compromise the semantic integrity in KGs as each relation is modeled separately in the relation-specific hyperplane or space. The general entity embeddings $\mathbf{h}$ and $\mathbf{t}$ are not explicitly translated by relation vectors. Although our model can also be viewed as a kind of translational KG embedding, we introduce the edge-centric model that contextualizes relations with edge embeddings.

Besides, there are some \emph{bilinear} models that exploit similarity-based functions to compute the energy of relational triples. DistMult~\cite{DistMult} and ComplEx~\cite{ComplEx} use the bilinear Hadamard product to compute energy. HolE~\cite{HolE} substitutes the Hadamard product with circular correlation. Analogy~\cite{Analogy} imposes analogical properties on embeddings. SimplE~\cite{SimplE} proposes an enhancement of Canonical Polyadic (CP) decomposition to compute energy. CrossE~\cite{CrossE} exploits to simulate the crossover interactions between entities and relations. RotatE~\cite{RotatE} defines each relation as a rotation from the head entity to the tail in the complex-valued embedding space. Recently, there are also some \emph{neural} embedding models including ProjE~\cite{ProjE}, ConvE~\cite{ConvE}, R-GCN~\cite{R-GCN} and ConvKB~\cite{ConvKB}. These bilinear and neural models achieve superior results on link prediction at the cost of much higher model complexity. Besides, many of these embedding models also have the identified shortcomings, such as HolE and ProjE. 

\subsection{KG Embeddings for Entity Alignment}
\label{sec:related_work_align}
Recently, several embedding-based entity alignment models have been proposed. MTransE~\cite{MTransE} captures two KG-specific vector spaces and jointly learns a transformation between them.
IPTransE \cite{IPTransE} employs PTransE \cite{PTransE} to embed two KGs into a unified vector space. It iteratively updates alignment information through a self-training technique. JAPE~\cite{JAPE} incorporates attribute embeddings for entity alignment. BootEA~\cite{BootEA} solves the entity alignment problem in a bootstrapping manner. KDCoE~\cite{KDCoE} co-trains description embeddings and structure embeddings to incorporate both the literal and structural information of KGs for entity alignment. GCN-Align~\cite{GCN-Align} employs graph convolutional networks to learn KG embeddings for entity alignment. AttrE~\cite{AttrE} regards literal values as ``virtual entities'' and uses TransE to embed the attribute triples for entity alignment. Note that, some of these models exploit additional resources in KGs for entity alignment, such as relation paths (IPTransE), textual descriptions (KDCoE) and literal values (AttrE). By contrast, the proposed TransEdge leverages the basic relational structures for KG embedding, without using additional resources.

\section{Edge-centric Knowledge Graph Embedding}
\label{sec:model}

\begin{figure}[!t]
	\setlength{\abovecaptionskip}{5pt}
	\setlength{\belowcaptionskip}{0pt}
	\centering
	\includegraphics[width=0.999\textwidth]{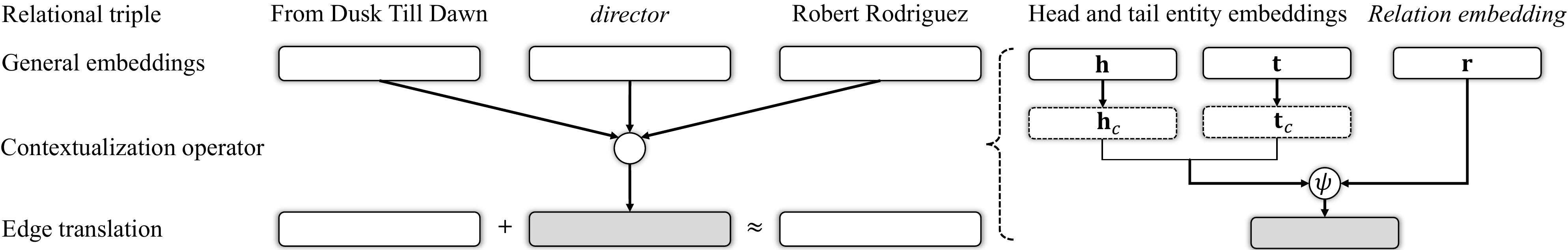}
	\caption{Illustration of the key idea of relation-contextualized KG embeddings. The white boxes denote the general embeddings of entities and relations, and the gray boxes denote the contextualized representation for this relation, i.e., the edge embedding. $\mathbf{h}_c$ and $\mathbf{t}_c$ are the interaction embeddings for entities. $\psi$ is a contextualization operator.}	
	\label{fig:rel_trans}
\end{figure}

TransEdge embeds the entities and relations of KGs in a $d$-dimensional vector space. Unlike the conventional relation-level models, for a relational triple, the head and tail entity embeddings in TransEdge hold an edge translation. Fig.~\ref{fig:rel_trans} illustrates the main idea. The contextualization operator $\psi$ takes as input the combined embeddings of the head and tail entities (edge direction) as well as the relation embedding (edge label) to compute edge embeddings.

\subsection{Formulation of Energy Function}
Like TransE, we define an energy function to measure the error of edge translation between entity embeddings. For simplicity, the energy of a relational triple $(h,r,t)$ in TransEdge is written as follows:
\begin{equation}\label{eq:score}
f(h,r,t)=||\mathbf{h} + \psi(\mathbf{h}_c,\mathbf{t}_c, \mathbf{r}) -\mathbf{t}||.
\end{equation}
The edge embedding $\psi(\mathbf{h}_c,\mathbf{t}_c, \mathbf{r})$ corresponds to a translation vector between the head to tail entity embeddings. In TransEdge, we learn a general embedding for each entity, such as $\mathbf{h}$ for $h$. General embeddings capture the geometric positions and relational semantics of entities in the vector space. We also introduce interaction embeddings for entities, such as $\mathbf{h}_c$ for $h$, which are used to encode their participation in the calculation of edge embeddings. Separating the interaction embeddings from general ones can avoid the interference of such two different information. 

\subsection{Contextualization Operation}
\begin{figure}[!t]
	\setlength{\abovecaptionskip}{5pt}
	\setlength{\belowcaptionskip}{0pt}
	\centering
	\includegraphics[width=0.9\textwidth]{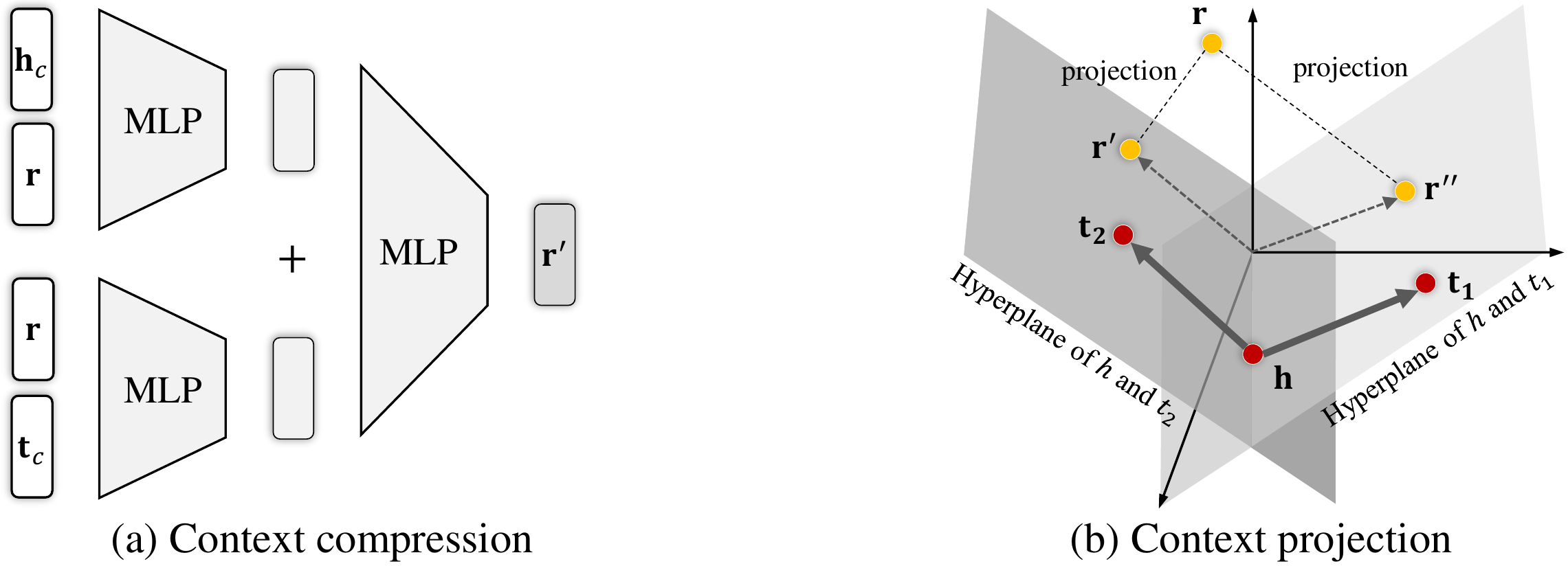}
	\caption{Illustration of the proposed contextualization operations.}	
	\label{fig:operation}
\end{figure}

The calculation of edge embeddings $\psi(\mathbf{h}_c,\mathbf{t}_c,\mathbf{r})$ should involve the information of both the head and tail entities (edge direction), as well as the relations (edge label). We study two different methods shown in Fig.~\ref{fig:operation}, which are discussed in detail below.

\subsubsection{Context Compression}
This method uses multilayer perceptrons (MLPs) to compress the embeddings of the edge direction and label. Specifically, given a MLP with one hidden layer (i.e., two layers in total plus the output layer) and the input vector $\mathbf{v}^{(0)}$, each layer is calculated with a set of weight matrices $\mathbf{W}$ and vectors $\mathbf{b}$:
\begin{equation} \label{eq:nn}
\small
\mathbf{v}^{(1)} = \sigma \big(\mathbf{W}^{(1)}\,\mathbf{v}^{(0)} + \mathbf{b}^{(1)}\big), \,\,\,
\mathbf{v}^{(2)} = \sigma \big(\mathbf{W}^{(2)}\, \mathbf{v}^{(1)} + \mathbf{b}^{(2)}\big), 
\end{equation}
where $\sigma() $ is the activation function like $\tanh()$. Finally, $\textsc{mlp}(\mathbf{v}^{(0)})=\mathbf{v}^{(2)}$. As illustrated in Fig.~\ref{fig:operation}(a), given a relational triple $(h,r,t)$, we concatenate $\mathbf{h}_c$ and $\mathbf{r}$ as input and feed it to a MLP to get a combined representation. $\mathbf{t}_c$ and $\mathbf{r}$ are encoded in the same way. Finally, we employ another MLP to combine them. The three MLPs capture the non-linear combination of the representations of edge direction and label. Let $\textsc{mlp}$() denote a MLP. The edge embedding is calculated as follows:
\begin{equation}\label{eq:sm}
\psi(\mathbf{h}_c,\mathbf{t}_c,\mathbf{r}) = \textsc{mlp}_1(\textsc{mlp}_2([\mathbf{h}_c;\mathbf{r}])+\textsc{mlp}_3([\mathbf{r};\mathbf{t}_c])),
\end{equation}
where $[\mathbf{h}_c;\mathbf{r}] = \textsc{concat}(\mathbf{h}_c,\mathbf{r}) \in \mathbb{R}^{2d} $, which concatenates the given vectors. 

\subsubsection{Context Projection}
Projecting embeddings onto hyperplanes~\cite{HyTE,TransH} has shown promising effects on the processing of disparate feature representations. Here, we regard the edge direction and label representations as orthogonal features and project the label representation onto the hyperplane of the edge direction representations, as illustrated in Fig.~\ref{fig:operation}(b). Given two relational triples $(h,r,t_1)$ and $(h,r,t_2)$, $\mathbf{r}'$ and $\mathbf{r}''$ are two edge embeddings for $\mathbf{r}$ projected on hyperplanes. Let $\mathbf{w}_{(h,t)}$ be the normal vector of such hyperplane. The edge embedding for ($h,r,t$) is calculated by vector projection as follows:
\begin{equation}\label{eq:bias}
\psi(\mathbf{h}_c,\mathbf{t}_c,\mathbf{r}) = \mathbf{r} - \mathbf{w}_{(h,t)}^\top\mathbf{r}\,\mathbf{w}_{(h,t)}.
\end{equation}
We use a MLP to compute the normal vector based on the concatenated embeddings of head and tail entities. Formally, $\mathbf{w}_{(h,t)}=\textsc{mlp}([\mathbf{h}_c; \mathbf{t}_c])$, s.t. $||\mathbf{w}_{(h,t)}||=1$.

\subsection{Loss Function}
Following the conventional training strategy of previous models~\cite{KGE_survey}, we train TransEdge based on the local-closed world assumption. In this case, we regard the observed relational triples in KGs as positive examples while the unobserved ones as negative samples (either false or missing triples). In our model, positive relational triples are expected to fulfill such relation-contextualized translation with low energy. Negative relational triples are supposed to hold higher energy as they are more invalid than positive ones. To this end, we minimize the following limit-based loss~\cite{BootEA}, which can create more distinguishable embedding structures than the conventional marginal ranking loss:
\begin{equation}\label{eq:rel_embed_obj}
\small
\mathcal{L} = \sum_{(h,r,t) \in \mathcal{T}}[f(h,r,t) - \gamma_1]_+ + \sum_{(h',r',t') \in \mathcal{T}^-}\alpha\,[\gamma_2 - f(h',r',t')]_+,
\end{equation}
where $[x]_+ = \max(0,x)$. $\gamma_1,\gamma_2$ are the hyper-parameters to control the energy of triples, s.t. $\gamma_1 < \gamma_2$. $\alpha$ is a hyper-parameter to balance the positive and negative samples. $ \mathcal{T}^- $ denotes the set of negative triples, which can be generated by some heuristic strategies. Here, we choose the truncated negative sampling~\cite{BootEA}, which generates negative triples by replacing either the head or tail entities of positive relational triples with some random neighbors of these entities. 

\subsection{Implementation for Entity Alignment}
Given a source KG $\mathcal{K}_1=(\mathcal{E}_1,\mathcal{R}_1,\mathcal{T}_1)$ and a target KG $\mathcal{K}_2=(\mathcal{E}_2,\mathcal{R}_2,\mathcal{T}_2)$, entity alignment seeks to find entities from different KGs that refer to the same real-world object. Embedding-based entity alignment helps overcome the semantic heterogeneity in different KGs and receives increasing attention recently.

For entity alignment, we let each entity pair in seed alignment (i.e., training data) share the same embedding (called parameter sharing), to reconcile $\mathcal{K}_1$ and $\mathcal{K}_2$. In this way, the two KGs are merged into one and we can use TransEdge to learn entity embeddings from this ``combined KG''. For training, semi-supervised strategies, such as self-training and co-training, have been widely used for embedding-based entity alignment~\cite{KDCoE,BootEA,IPTransE}. This is because the size of seed alignment is usually small. For example, as investigated in~\cite{MTransE}, in Wikipedia, the inter-lingual links cover less than 15\% entity alignment. To cope with this problem, we use the bootstrapping strategy \cite{BootEA} to iteratively select the likely-aligned entity pairs, which we denote by $\mathcal{D}=\{(e_1, e_2)\in\mathcal{E}_1 \times\mathcal{E}_2 |\cos(\mathbf{e}_1, \mathbf{e}_2) > s\}$, where $s$ is the similarity threshold. As errors in the newly-found entity alignment are unavoidable, we do not make each newly-found entity pair share the same embedding. Instead, we minimize the following loss to let the proposed entity alignment has a small embedding distance (i.e., high similarity):
\begin{equation}\label{eq:semi}
\mathcal{L}_{\text{semi}} = \sum_{(e_1,e_2) \in \mathcal{D}}||\mathbf{e}_1 - \mathbf{e}_2||.
\end{equation}
In the test phase, given an entity to be aligned in $\mathcal{K}_1$, we rank entities in $\mathcal{K}_2$ as its counterpart candidates in descending order based on the cosine similarity of entity embeddings. The right counterpart is expected to have a top rank.

The parameters of TransEdge are initialized using the Xavier initializer~\cite{Xavier}. The embedding loss $\mathcal{L}$ on $\mathcal{T}_1 \cup \mathcal{T}_2$ and the semi-supervised training loss $\mathcal{L}_{\text{semi}}$ are jointly optimized using a stochastic gradient descent algorithm AdaGrad. We enforce the $L_2$ norm of KG embeddings to 1 to reduce the trivial learning by artificially increasing the embedding norms~\cite{TransE}. The variants of TransEdge that use context compression (CC) and context projection (CP) are denoted by TransEdge-CC and TransEdge-CP, respectively. For ablation study, we also develop the degraded variants of TransEdge without using semi-supervised training, which are marked by the suffix (w/o semi).

\subsection{Implementation for Link Prediction}
Link prediction is the task of inferring the missing head or tail entities when given incomplete relational triples. For example, given ($\rule{0.38cm}{0.15mm}$, \textit{capitalOf}, New Zealand), the link prediction models are expected to rank the right head entity Wellington at the first place. Link prediction is a key task for KG completion and has been widely used as an evaluation task by many previous KG embedding models.

The embeddings are learned by minimizing $\mathcal{L}$. The parameters are initialized using the Xavier initializer and the loss is also optimized using AdaGrad. In the test phrase, for head prediction $(\rule{0.38cm}{0.15mm},r,t)$, we create a set of candidate triples by replacing $\rule{0.38cm}{0.15mm}$ with all possible entities. The candidate triples can be ranked in ascending order according to their energy calculated using Eq.~(\ref{eq:score}). The right candidate triple is expected to have a top rank. Tail prediction $(h,r,\rule{0.38cm}{0.15mm})$ can be done in the same way.

\subsection{Complexity Analysis}
\label{subsect:complexity}
In general, TransEdge learns two embeddings for each entity. We provide a complexity comparison in Table \ref{tab:complexity}, where $n_e$ and $n_r$ denote the numbers of entities and relations, respectively, and $d$ is the embedding dimension. As our model introduces additional parameters for embedding entities. its complexity is $O(2n_ed + n_rd)$, which is more than that of TransE. However, it is less than the complexity of TransD. Note that, the parameter complexity of TransEdge grows linearly with the number of entities and the embedding dimension.

\begin{table}[!t]
	\setlength{\abovecaptionskip}{0pt}
	\setlength{\belowcaptionskip}{5pt}	
	\caption{Complexity comparison of translational embedding models}
	\label{tab:complexity}
	\centering\small
	\begin{tabular}{l|l}
		\toprule 
			\ Model &\ \#Embeddings \\ \hline
			\ TransE \cite{TransE} & \ $O(n_ed + n_rd)$ \\
			\ TransH \cite{TransH} & \ $O(n_ed + 2n_rd)$ \\
			\ TransR \cite{TransR} & \ $O(n_ed + n_rd^2)$ \\
			\ TransD \cite{TransD} & \ $O(2n_ed + 2n_rd)$ \quad\\
		\hline
			\ TransEdge (this paper) \quad & \ $O(2n_ed + n_rd)$ \\
		\bottomrule
	\end{tabular}
\end{table}

\section{Experiments}
\label{sec:exp}
We assess TransEdge on two popular embedding-based tasks: entity alignment between two KGs and link prediction in one KG. The source code of TransEdge is available online\footnote{\url{https://github.com/nju-websoft/TransEdge}}.

\subsection{Task 1: Embedding-based Entity Alignment}

\subsubsection{Datasets.}
To evaluate TransEdge on various scenarios of entity alignment, we choose the following datasets: (1) DBP15K~\cite{JAPE} is extracted from the multilingual DBpedia. It contains three cross-lingual entity alignment datasets: DBP\textsubscript{ZH-EN} (Chinese to English), DBP\textsubscript{JA-EN} (Japanese to English) and DBP\textsubscript{FR-EN} (French to English). Each dataset has 15 thousand aligned entity pairs. (2) DWY100K~\cite{BootEA} has two large-scale monolingual datasets, DBP-WD and DBP-YG, sampled from DBpedia, Wikidata and YAGO3. Each dataset has 100 thousand aligned entity pairs. For a fair comparison, we reuse their original dataset splits in evaluation.

\subsubsection{Competitive Models.}
For comparison, we choose the following state-of-the-art embedding-based entity alignment models: MTransE~\cite{MTransE}, IPTransE~\cite{IPTransE}, JAPE~\cite{JAPE}, BootEA~\cite{BootEA} and its non-bootstrapping version AlignE, as well as GCN-Align~\cite{GCN-Align}. We do not compare with some other models like KDCoE~\cite{KDCoE} and AttrE~\cite{AttrE}, since they require additional resources (e.g., textual descriptions and attribute values) that do not present in our problem setting as well as other competitors'. Furthermore, the character-based literal embedding used in AttrE~\cite{AttrE} is unsuited to cross-lingual entity alignment as the characters of different languages (such as Chinese and English) can be very heterogeneous. Our goal is to exploit the basic relational structures of KGs for entity alignment.

To further understand the benefits and limitations of KG embeddings for entity alignment, we extend several representative embedding models that are used for link prediction as the competitors, including: three translational models TransH~\cite{TransH}, TransR~\cite{TransR} and TransD~\cite{TransD}; two bilinear models HolE~\cite{HolE} and SimplE~\cite{SimplE}; and two neural models ProjE~\cite{ProjE} and ConvE~\cite{ConvE}. Note that ComplEx~\cite{ComplEx} is very similar to HolE~\cite{RotatE}. So, we pick HolE as the representative. We do not include Analogy~\cite{Analogy} and ConvKB~\cite{ConvKB}, because we find that these methods do not perform well on the datasets. Similar to TransEdge, we merge two KGs into one via parameter sharing and use these models to learn embeddings. We refer the open-source KG embedding framework OpenKE~\cite{OpenKE} to implement TransH, TransR, TransD and HolE, while SimplE, ProjE and ConvE are implemented based on their code.

\subsubsection{Experimental Settings.}
We have tuned a series of hyper-parameters. For example, we select the learning rate among \{0.001, 0.005, 0.01, 0.02\} and the positive margin $\gamma_1$ among \{0.1, 0.2, $\cdots$, 0.5\}. The selected setting of hyper-parameters is reported as follows. For TransEdge-CC, $\gamma_1 = 0.3$, $\gamma_2 = 2.0$, $\alpha = 0.3$, $s = 0.75$, $d = 75$. For TransEdge-CP, $\gamma_1 = 0.2$, $\gamma_2 = 2.0$, $\alpha = 0.8$, $s = 0.7$, $d = 75$. The activation function is $\tanh()$ for MLPs. For DBP15K, we generate $20$ negative samples for each relational triple and the batch size is $2,000$. For DWY100K, we generate $25$ negative samples for each relational triple and the batch size is $20,000$. We adopt $L_2$-norm in the energy function. The learning rate is $0.01$ and the training is terminated using early stop based on the Hits@1 performance to avoid overfitting. We use CSLS \cite{Word_Translation} as similarity measure. We choose three widely-used metrics: Hits$@k$, mean rank (MR) and mean reciprocal rank (MRR). Higher Hits$@k$ and MRR values, and lower MR values indicate better performance. Note that, Hits@1 is equivalent to precision, and MRR is more robust than MR since MRR is more able to tolerate a few poorly-ranked correct candidates.

\subsubsection{Entity Alignment Results.}
\begin{table}[!t]
	\setlength{\abovecaptionskip}{0pt}
	\setlength{\belowcaptionskip}{5pt}
	\centering
	\caption{Entity alignment results on DBP15K}
	\label{tab:ent_alignment}
	\resizebox{1.0\textwidth}{!}{
		\begin{threeparttable}\scriptsize
			\setlength{\tabcolsep}{0.3em}
			\begin{tabular}{lcccccccccccc}
				\toprule
				& \multicolumn{4}{c}{DBP\textsubscript{ZH-EN}} & \multicolumn{4}{c}{DBP\textsubscript{JA-EN}} & \multicolumn{4}{c}{DBP\textsubscript{FR-EN}} \\
				\cmidrule(lr){2-5} \cmidrule(lr){6-9} \cmidrule(lr){10-13} 
				& Hits@1 & Hits@10 & MRR & MR & Hits@1 & Hits@10 & MRR & MR & Hits@1 & Hits@10 & MRR & MR \\ 
				\midrule
				MTransE~\cite{MTransE}~$^\dag$ & 0.308 & 0.614 & 0.364 & 154 & 0.279 & 0.575 & 0.349 &159 & 0.244 & 0.556 & 0.335 &139  \\
				IPTransE~\cite{IPTransE}~$^\ddag$ & 0.406 & 0.735 & 0.516& --  & 0.367 & 0.693 & 0.474& --  & 0.333 & 0.685 & 0.451& --  \\
				JAPE~\cite{JAPE}~$^\dag$ & 0.412 & 0.745 & 0.490& 64  & 0.363 & 0.685 & 0.476& 99  & 0.324 & 0.667& 0.430& 92   \\  
				AlignE~\cite{BootEA} & 0.472 & 0.792 & 0.581& --  & 0.448  & 0.789 & 0.563& --  & 0.481 & 0.824 & 0.599& --   \\
				BootEA~\cite{BootEA} & 0.629 & 0.848 & 0.703& --  & 0.622 & 0.854 & 0.701& --  & 0.653 & 0.874 & 0.731& --  \\	
				GCN-Align~\cite{GCN-Align} &  0.413 & 0.744 & -- & -- & 0.399 & 0.745 & -- & -- & 0.373 & 0.745 & --& --   \\
				\midrule
				TransH~\cite{TransH}~$^\triangle$ & 0.377 & 0.711 & 0.490& 52  & 0.339 & 0.681 & 0.462& 59  & 0.313 & 0.668 & 0.433& 47  \\
				TransR~\cite{TransR}~$^\triangle$ & 0.259 & 0.529 & 0.349& 299  & 0.222 & 0.440 & 0.295& 315  & 0.059 & 0.225 & 0.116& 502  \\
				TransD~\cite{TransD}~$^\triangle$ & 0.392 & 0.729 & 0.505& 48  & 0.356 & 0.695 & 0.468& 58  & 0.323 & 0.694 & 0.447& 43  \\
				\midrule
				HolE~\cite{HolE}~$^\triangle$ & 0.250 & 0.535 & 0.346& 488  & 0.256 & 0.517 & 0.343& 560  & 0.149 & 0.465 & 0.251& 1133   \\
				SimplE~\cite{SimplE}~$^\diamondsuit$ &0.317 & 0.575 &0.405 & 453 &0.255 & 0.525& 0.346 & 409 &0.147&0.438&0.241& 397  \\
				\midrule
				ProjE~\cite{ProjE}~$^\diamondsuit$ & 0.290 & 0.527 & 0.374& 705  & 0.273 & 0.475 & 0.345& 919  & 0.283 & 0.527 & 0.368& 659  \\
				ConvE~\cite{ConvE}~$^\diamondsuit$ & 0.169 & 0.329 & 0.224& 1123  & 0.192 & 0.343 & 0.246& 1081  & 0.240 & 0.459 & 0.316& 694  \\
				\midrule
				TransEdge-CC (w/o semi)  &  0.622 & 0.868 & 0.711 & 65 & 0.601 & 0.863 & 0.696 & 56 & 0.617 & 0.891 & 0.716 & 38\\
				TransEdge-CP (w/o semi)  & 0.659 & 0.903 & 0.748 & 50 & 0.646 & 0.907 & 0.741 & 36 & 0.649 & 0.921 & 0.746 & 25\\
				TransEdge-CC  & 0.669 & 0.871 & 0.744 & 66 & 0.645 & 0.859 & 0.722 & 67 & 0.666 & 0.893 & 0.749 & 40 \\
				TransEdge-CP  & \textbf{0.735} & \textbf{0.919} & \textbf{0.801} & \textbf{32} & \textbf{0.719} & \textbf{0.932} & \textbf{0.795} & \textbf{25} & \textbf{0.710} & \textbf{0.941} & \textbf{0.796} & \textbf{12} \\
				\bottomrule
			\end{tabular}
			$\dag$ Hits@$k$ and MR results are taken from~\cite{JAPE} while MRR results are taken from~\cite{BootEA}. $\ddag$ Results are taken from~\cite{BootEA}. $\triangle$ Results are produced by ourselves using OpenKE~\cite{OpenKE}. $\diamondsuit$ Results are produced by ourselves using their source code. $-$ denotes unreported results in their papers. Unmarked results are taken from their own papers. Best results are marked in boldface, and same in the following tables.
	\end{threeparttable}}
\end{table}

The results of entity alignment are depicted in Tables~\ref{tab:ent_alignment} and~\ref{tab:ent_alignment_100k}. We can see that TransEdge consistently achieves the best for all the metrics on the five datasets. For example, on DBP\textsubscript{ZH-EN}, TransEdge-CP (w/o semi) achieves an improvement of $ 0.187$ on Hits@$1$ against AlignE. If compared with its bootstrapping version BootEA, TransEdge-CP (w/o semi) still achieves a gain of $0.030$ while the improvement of TransEdge-CP reaches $0.106$. We can see that BootEA is a very competitive model due to its powerful bootstrapping strategy. However, our semi-supervised variants TransEdge-CC and TransEdge-CP significantly outperform BootEA on DBP15K. This is due to the ability of TransEdge on preserving KG structures. 

On DBP15K, both TransEdge-CC and TransEdge-CP show good performance. TransEdge-CC (w/o semi) still obtains superior results than AlignE and TransEdge-CC also outperforms BootEA. Furthermore, we find that TransEdge-CP achieves better results than TransEdge-CC. We think that this is because the context projection has a good geometric interpretation, as shown in Fig.~\ref{fig:rel_trans}(b), which helps capture better and more solid relational structures of KGs for entity alignment. We can also see that the proposed semi-supervised training for entity alignment brings remarkable improvement. For example, on DBP\textsubscript{ZH-EN}, it increases the Hits@1 scores of TransEdge-CP from $0.659$ (w/o semi) to $0.735$. These results indicate that the proposed context compression and projection can both accurately compute the edge embeddings. The proposed semi-supervised training also contributes to the performance improvement.

\begin{table}[!t]
	\setlength{\abovecaptionskip}{0pt}
	\setlength{\belowcaptionskip}{5pt}
	\centering
	\caption{Entity alignment results on DWY100K}
	\label{tab:ent_alignment_100k}
	\resizebox{0.85\textwidth}{!}{
		\begin{threeparttable}\scriptsize
			\setlength{\tabcolsep}{0.3em}
			\begin{tabular}{lcccccccc}
				\toprule
				& \multicolumn{4}{c}{DBP-WD} & \multicolumn{4}{c}{DBP-YG} \\
				\cmidrule(lr){2-5} \cmidrule(lr){6-9} 
				& Hits@1 & Hits@10 & MRR & MR & Hits@1 & Hits@10 & MRR & MR \\ 
				\midrule
				MTransE~\cite{MTransE}~$^\ddag$ & 0.281 & 0.520 & 0.363 & -- & 0.252 & 0.493 & 0.334 & -- \\
				IPTransE~\cite{IPTransE}~$^\ddag$  & 0.349 & 0.638 & 0.447 & --& 0.297 & 0.558 & 0.386 & -- \\
				JAPE~\cite{JAPE}~$^\ddag$  & 0.318 & 0.589 & 0.411& -- & 0.236 & 0.484 & 0.320& --\\  
				AlignE~\cite{BootEA}~$^\ddag$ & 0.566 & 0.827 & 0.655& --  & 0.633 & 0.848 & 0.707& --  \\
				BootEA~\cite{BootEA}~$^\ddag$ & 0.748 & 0.898 & 0.801& -- & 0.761 & 0.894 & 0.808& -- \\	
				GCN-Align~\cite{GCN-Align}~$^\nabla$ & 0.479 & 0.760 & 0.578& 1988 & 0.601 & 0.841 & 0.686& 299  \\
				\midrule
				TransH~\cite{TransH}~$^\triangle$ & 0.351 & 0.641 & 0.450 & 117 & 0.314 & 0.574 & 0.402& 90  \\
				TransR~\cite{TransR}~$^\triangle$ & 0.013 & 0.062 & 0.031& 2773  & 0.010 & 0.052 & 0.026& 2852 \\
				TransD~\cite{TransD}~$^\triangle$ & 0.362 & 0.651 & 0.456& 152  & 0.335 & 0.597 & 0.421 & 90\\
				\midrule
				HolE~\cite{HolE}~$^\triangle$& 0.223 & 0.452 & 0.289& 811  & 0.250 & 0.484 & 0.327 & 437 \\
				SimplE~\cite{SimplE}~$^\diamondsuit$& 0.169 & 0.328 & 0.223 & 3278 & 0.131 & 0.282 & 0.183 & 3282\\
				\midrule
				ProjE~\cite{ProjE}~$^\diamondsuit$ & 0.312 & 0.504 & 0.382& 2518  & 0.366 & 0.573 & 0.436& 1672  \\
				ConvE~\cite{ConvE}~$^\diamondsuit$ & 0.403 & 0.628 & 0.483& 1428  & 0.503 & 0.736 & 0.582& 837 \\
				
				\midrule
				TransEdge-CC (w/o semi)  & 0.687 & 0.910 & 0.767 & 70 & 0.759 & 0.935 & 0.822 & 24 \\
				TransEdge-CP (w/o semi)  & 0.692 & 0.898 & 0.770 & 106 & 0.726 & 0.909 & 0.792 & 46 \\
				TransEdge-CC  & 0.732 & 0.926 & 0.803 & \textbf{65} & 0.784 & \textbf{0.948} & \textbf{0.844} & \textbf{22} \\
				TransEdge-CP  & \textbf{0.788} & \textbf{0.938} & \textbf{0.824} & 72 & \textbf{0.792} & 0.936 & 0.832 & 43\\
				\bottomrule
			\end{tabular}
			$\nabla$: Results are produced using its code. Other marks mean the same in Table~\ref{tab:ent_alignment}.
		\end{threeparttable}
	}
\end{table}

We notice that, on DWY100K, the improvement of TransEdge is not so large as that on DBP15K. For example, on DBP-WD, TransEdge-CP only achieves an improvement of $ 0.040$ on Hits@$1$ against BootEA. We think this is because the two KGs in DBP-WD or DBP-YG have aligned relational structures and their entities are one to one aligned. But in DBP15K, there are many noisy entities that have no counterparts. Thus, DWY100K is relatively simple for entity alignment. On datasets with noisy entities, TransEdge gives a big advantage to others, which indicates the robustness of TransEdge.

It is interesting to see that some modified models also demonstrate competitive performance on entity alignment. ConvE even outperforms some alignment-oriented embedding models such as MTransE, IPTransE and JAPE on the DWY100K datasets, which indicates the potential of deep learning techniques. We also notice that the performance of TransR is very unstable. It achieves promising results on DBP\textsubscript{ZH-EN} and DBP\textsubscript{JA-EN} but fails on the other three datasets. We take a closer look at the five datasets and discover that DBP\textsubscript{ZH-EN} and DBP\textsubscript{JA-EN} contain some relation alignment. When TransR performs relation-specific projections on entities, the relation alignment would pass some alignment information to entities. The requirement of relation alignment limits the applicability of TransR to entity alignment. We can conclude that not all embedding models designed for link prediction are suitable for entity alignment. 

\subsection{Task 2: Embedding-based Link Prediction}
\label{sect:link}

\subsubsection{Datasets.} We use two benchmark datasets FB15K-237~\cite{FB15k237} and WN18RR~\cite{ConvE} for link prediction. They are the improved versions of FB15K~\cite{TransE} and WN18~\cite{TransE}, respectively. As found in~\cite{ConvE,FB15k237}, FB15K and WN18 contain many symmetric triples that are easy to infer by learning some trivial patterns. Thus, the work in~\cite{ConvE,FB15k237} creates FB15K-237 and WN18RR by removing inverse relations from the testing sets. So, FB15K-237 and WN18RR are more difficult, and both of them have gradually become the most popular benchmark datasets for link prediction in recent studies~\cite{ConvE,ConvKB,RotatE,CrossE}. FB15K-237 contains $14,541$ entities, $237$ relations and $310,116$ relational triples. WN18RR has $40,943$ entities, $11$ relations and $93,003$ relational triples. For a fair comparison, we reuse the original training/validation/test splits of the two datasets in evaluation.

\subsubsection{Competitive Models.}
For comparison, we choose a wide range of embedding models for link prediction as the competitors, including five translational models, seven bilinear models and five neural models, as listed in Table \ref{tab:link_pred_fb15k-237}. For the sake of fairness and objectivity, we report the published results of them as many as possible. But there still exist some results unavailable in the reference papers. If some models have not been evaluated on FB15K-237 or WN18RR, we use their released code to produce the results by ourselves. 

\subsubsection{Experimental Settings.} 
We have tuned hyper-parameter values by a careful grid search. The selected setting for hyper-parameters is as follows. For FB15K-237, $\gamma_1 = 0.4$, $\gamma_2 = 0.9$, $\alpha = 0.4$, $d = 200$. The batch size is 200 and the learning rate is 0.005. We generate 10 negative samples for each triple. For WN18RR, $\gamma_1 = 0.2$, $\gamma_2 = 2.7$, $\alpha = 0.8$, $d = 500$. The batch size is $2,000$ and the learning rate is 0.01. We sample 30 negatives for each triple. The activation function is still $\tanh()$ for MLPs. We use $L_2$-norm in our energy function. When evaluating the ranking lists, we use the filtered setting~\cite{TransE}, i.e., given a candidate triple list, we remove from the list all other positive triples that appear in the training/validation/test data. Then, we get a new filtered ranking list and the right triple is expected to have a high rank. By convention, we report the average results of head prediction and tail prediction. Same as embedding-based entity alignment, we use Hits@$k$, MR and MRR.

\begin{table}[!t]
	\setlength{\abovecaptionskip}{0pt}
	\setlength{\belowcaptionskip}{5pt}
	\centering
	\caption{Link prediction results on FB15K-237 and WN18RR}
	\label{tab:link_pred_fb15k-237}
	\resizebox{1.0\textwidth}{!}{
		\scriptsize
		\begin{threeparttable}
			\setlength{\tabcolsep}{0.5em}
			\begin{tabular}{llcccccccc}
				\toprule
				& & \multicolumn{4}{c}{FB15K-237} & \multicolumn{4}{c}{WN18RR}\\
				\cmidrule(lr){3-6} \cmidrule(lr){7-10}
				Model& Type & Hits@1 & Hits@10 & MRR & MR & Hits@1 & Hits@10 & MRR & MR \\ 
				\midrule
				TransE~\cite{TransE}~$^\dag$& Trans. & -- & 0.436 & 0.269& 285 & -- & 0.453 & 0.412 & 5429 \\
				TransH~\cite{TransH}~$^\dag$ & Trans. & -- & 0.453 & 0.281 & 292 & -- & 0.429 & 0.435 & 5102 \\
				TransR~\cite{TransR}~$^{\ddag \nabla}$ & Trans. & -- & 0.429 & 0.162 & 337 & 0.017 & 0.257 & 0.094 & 3708 \\
				TransD~\cite{TransD}~$^{\ddag \nabla}$ & Trans. & -- & 0.428 & 0.162 & 305 & 0.015 & 0.139 & 0.060 & 6644 \\
				PTransE~\cite{PTransE}~$^\triangle$ & Trans. & 0.210 & 0.501 & 0.314 & 299& 0.272 & 0.424 & 0.337 & 5686 \\
				\midrule
				DistMult~\cite{DistMult}~$^\S$& Bilinear & 0.155 & 0.419 & 0.241 & 254 & 0.390 & 0.490 & 0.430 & 5110 \\
				HolE~\cite{HolE}~$^{\natural \nabla}$& Bilinear & 0.133 & 0.391 & 0.222& -- & 0.284 & 0.346 & 0.308 & 4874 \\
				ComplEx~\cite{ComplEx}~$^\S$& Bilinear & 0.158 & 0.428 & 0.247 & 339 & 0.410 & 0.510 & 0.440 &5261 \\
				Analogy~\cite{Analogy}~$^{\sharp \nabla}$& Bilinear & 0.131 & 0.405 & 0.219 & -- & 0.389 & 0.441 & 0.407 & 3836 \\
				ProjE~\cite{ProjE}& Neural & -- & 0.461 & 0.294 & 246 & -- & 0.474 & 0.453 & 4407 \\
				ConvE~\cite{ConvE}& Neural & 0.239 & 0.491 & 0.316 & 246 & 0.390 & 0.480 & 0.460 & 5277 \\
				R-GCN~\cite{R-GCN}& Neural & 0.153 & 0.414 & 0.248 & -- & -- & -- & -- & -- \\
				ConvKB~\cite{ConvKB}& Neural & -- & 0.517 & \textbf{0.396}  & 257 & -- & 0.525 & 0.248 & 2554 \\
				CACL~\cite{CACL}& Neural & -- & 0.487 & 0.349  & 235 & -- & 0.543 & 0.472 & 3154 \\
				SimplE~\cite{SimplE}~$^\square$& Bilinear & 0.225 & 0.461 & 0.230 & -- & -- & -- & -- & -- \\
				CrossE~\cite{CrossE}~$^\diamondsuit$ & Bilinear & 0.211 & 0.474 & 0.299 &-- & 0.373 & 0.394 & 0.374 & 6091 \\
				RotatE~\cite{RotatE}  & Bilinear & 0.241 & \textbf{0.533} & 0.338  & \textbf{177} & 0.428 & \textbf{0.571} & \textbf{0.476} & 3340 \\
				\midrule
				TransEdge-CC & Trans. & 0.227 & 0.482 & 0.310 & 305  & 0.411 & 0.516 & 0.439 & \textbf{2452}\\
				TransEdge-CP & Trans. & \textbf{0.243} & 0.512 & 0.333 & 219 & \textbf{0.433} & 0.487 & 0.451 & 4866 \\
				\bottomrule
			\end{tabular}
			$\dag$: Results are taken from~\cite{CACL}. $\ddag$: Results of FB15K-237 are taken from~\cite{Re-eval}. $\nabla$: Results on WN18RR are produced using OpenKE~\cite{OpenKE}. $\triangle$: Results are produced using its source code. $\S$: Results are taken from~\cite{ConvE}. $\natural$: Results are taken from~\cite{R-GCN}. $\sharp$: Results are taken from~\cite{CrossE}.  $\square$: Results are produced using the published source code. We do not include its results of WN18RR because we find them not promising. $\diamondsuit$: Results of WN18RR are produced using its source code.
		\end{threeparttable}
	}
\end{table}
\subsubsection{Link Prediction Results.}
Table~\ref{tab:link_pred_fb15k-237} gives the link prediction results on FB15K-237 and WN18RR. We can see that TransEdge significantly outperforms the translational models TransE, TransH, TransR and PTransE. This is because the proposed edge-centric translation can distinguish the different contexts of relations, while the relation translation of the aforementioned models usually leads to indistinguishable relation embeddings when modeling complex relational structures. When compared with the bilinear and neural models, especially with the very latest model RotatE~\cite{RotatE}, TransEdge-CP still achieves the best Hits@1 scores on both datasets. The best Hits@1 performance shows that TransEdge-CP can precisely capture the relational structures of KGs for link prediction, rather than put all possible candidates with similar and ambiguous ranks. We can also see that TransEdge-CC obtains the best MR result on WN18RR. Considering that WN18RR only has 11 relations but $40,943$ entities, we think this is because the MLPs can well fit such complex relational structures. Although the scores of TransEdge by other metrics such as Hits@10 and MRR fall behind ConvKB and RotatE, the model complexity of TransEdge is lower than them. For example, the convolution operation of ConvE and ConvKB is more complicated than the matrix multiplication used in the MLPs of TransEdge. Besides, the Euclidean vector space of real numbers generated by TransEdge is simpler than the complex vector space of ComplEx and RotatE.

\begin{table}[!t]
	\setlength{\abovecaptionskip}{0pt}
	\setlength{\belowcaptionskip}{5pt}
	\centering
	\caption{Entity alignment results on DBP15K with double relations}
	\label{tab:ent_alignment_double}
	\resizebox{1.0\textwidth}{!}{
		\begin{threeparttable}\scriptsize
			\setlength{\tabcolsep}{0.3em}
			\begin{tabular}{lcccccc}
				\toprule
				& \multicolumn{2}{c}{DBP\textsubscript{ZH-EN} (double)} & \multicolumn{2}{c}{DBP\textsubscript{JA-EN} (double)} & \multicolumn{2}{c}{DBP\textsubscript{FR-EN} (double)} \\
				\cmidrule(lr){2-3} \cmidrule(lr){4-5} \cmidrule(lr){6-7} 
				& Hits@1 & Hits@1$\downarrow$ & Hits@1 & Hits@1$\downarrow$ & Hits@1 & Hits@1$\downarrow$ \\ 
				\midrule
				MTransE~\cite{MTransE} &  0.230 & 25.32\% & 0.232 & 16.85\% & 0.208 & 14.75\%\\
				\midrule
				TransEdge-CC (w/o semi) & 0.601 & 3.38\% & 0.578 & 3.82\% & 0.585 & 5.18\% \\
				TransEdge-CP (w/o semi) & \textbf{0.652} & \textbf{1.06}\% & \textbf{0.623} & \textbf{3.56}\% & \textbf{0.641} & \textbf{1.23}\% \\
				\bottomrule
			\end{tabular}
	\end{threeparttable}}
\end{table}

\subsection{Analysis on Complex Relational Structures in KGs}

\subsubsection{One Entity Pair with Multiple Relations.} 
For further comparison, we evaluate TransEdge on KGs with double relations. We create a dummy relation $r'$ for each relation $r$ and add a dummy triple $(h,r',t)$ for $(h,r,t)$. The dummy relations and triples would not change the relational structures of KGs, but they would exacerbate the effects of the cases of one entity pair with multiple relations. We compare TransEdge (w/o semi) with the relation-level translational model MTransE~\cite{MTransE}. Due to space limitation, we report the Hits@1 results on DBP15K and the decrease rates (marked as Hits@1$\downarrow$) compared with their performance in Table~\ref{tab:ent_alignment}. The results are listed in Table~\ref{tab:ent_alignment_double}. We can see that the performance of TransEdge shows less variation than MTransE. This indicates that the complex relational structures can indeed hinder entity alignment performance while TransEdge has superior performance on modeling such structures.

\subsubsection{Multiple Entity Pairs with one Relation.}
\begin{figure}[!t]
	\centering
	\setlength{\abovecaptionskip}{2pt}
	\setlength{\belowcaptionskip}{0pt}
	\includegraphics[width=0.85\textwidth]{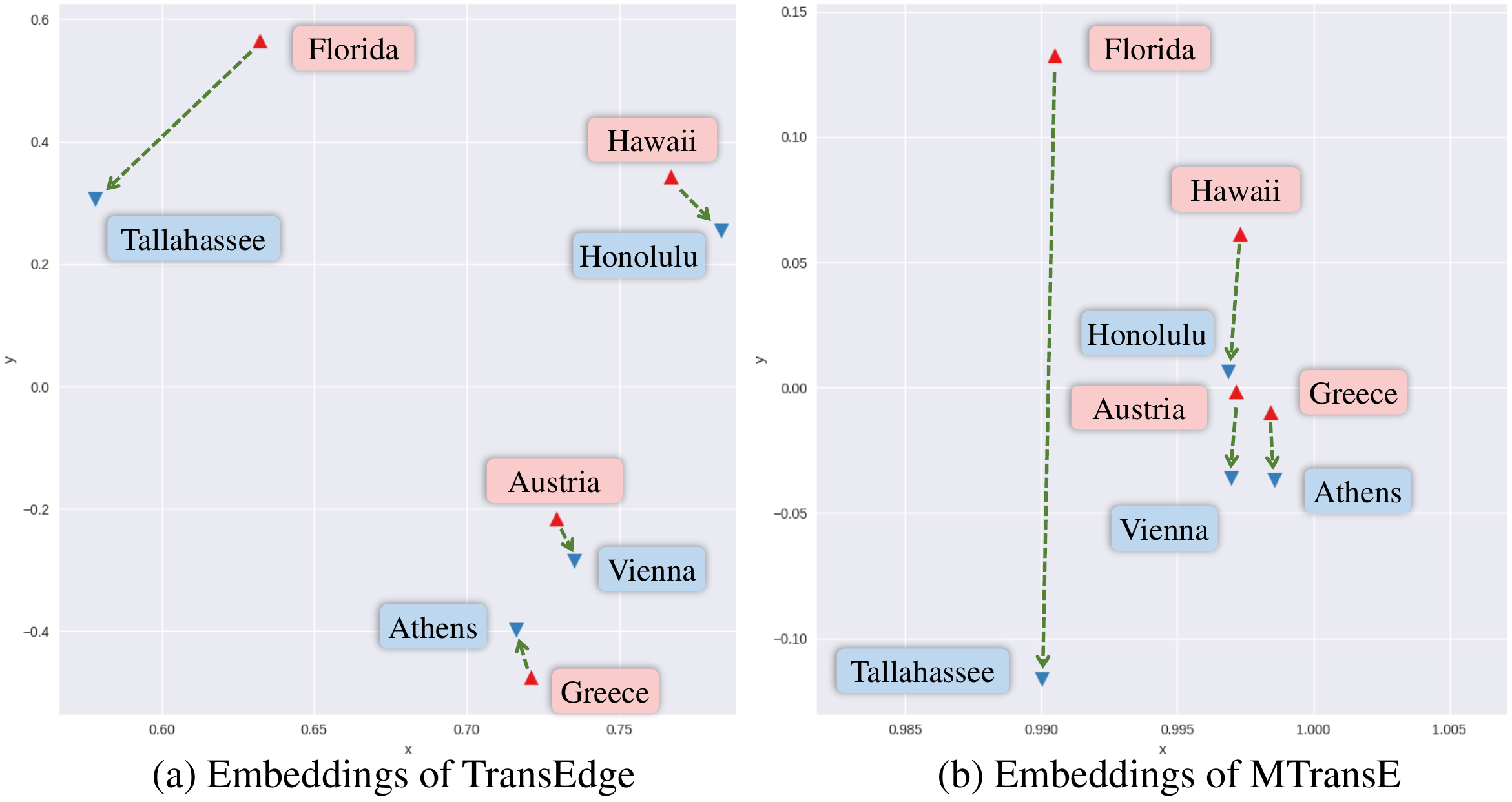}
	\caption{2D embedding projection of some countries (or states) and their \textit{capital} cities. The green arrows denote the translation vectors between entities.}
	\label{fig:viz}
\end{figure}

Figure~\ref{fig:viz} shows the 2D visualization for the embeddings of some entity pairs with the same relation \textit{capital} in DBP-WD. We project these embeddings into two dimensions using PCA. We can see that the embeddings of TransEdge show flexible and robust relational structures. The translation vectors of \textit{capital} are different in directions when involved in different contexts. For the embeddings of MTransE, the translation vectors are almost parallel. This means that, if several entities get involved in the same relational triple, they would be embedded very similarly by the same relational translation, which hinders the entity alignment performance. This experiment bears out the intuition of TransEdge illustrated by Fig.~\ref{fig:example}.

\begin{figure}[!t]
	\centering
	\setlength{\abovecaptionskip}{2pt}
	\setlength{\belowcaptionskip}{0pt}
	\includegraphics[width=0.9\textwidth]{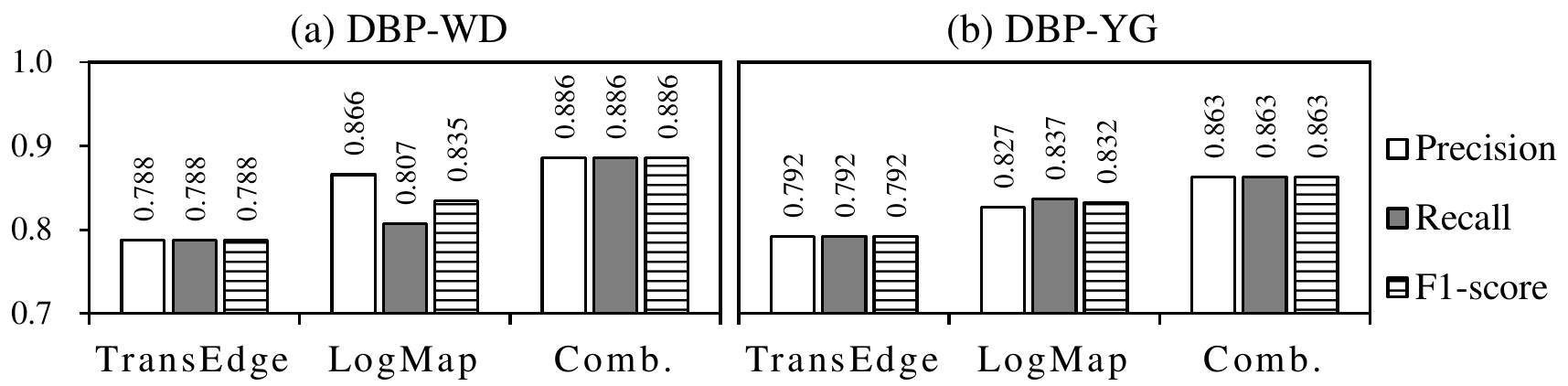}
	\caption{Results of TransEdge, LogMap~\cite{LogMap} and their combination on DWY100K.}
	\label{fig:logmap}
\end{figure}

\subsection{Comparison with Conventional Entity Alignment Method} 
Conventional entity alignment methods usually exploit literal attributes like names and comments, or OWL logics, to identify similar entities, which are quite different from TransEdge. We further compare TransEdge-CP with LogMap~\cite{LogMap}, a popular and accessible conventional entity alignment method. We use its web front-end system\footnote{\url{http://krrwebtools.cs.ox.ac.uk/logmap/}} to obtain its performance on the monolingual datasets DBP-WD and DBP-YG. We also design a strategy to combine TransEdge-CP and LogMap, which combines their produced entity alignment (i.e., Hits@1 alignment for TransEdge) by voting based on the predicted similarity. 
We report the conventional precision, recall and F1-score results in Fig.~\ref{fig:logmap}. Note that, for embedding-based entity alignment, recall and F1-score are equal to precision, because we can always get a candidate list for each input entity based on their embeddings. We can see that LogMap shows very competitive performance and it outperforms TransEdge and the other embedding-based models. However, we find that the combined results achieve the best. This shows that TransEdge is complementary with conventional entity alignment methods.

\section{Conclusion and Future Work}
\label{sec:conc}
In this paper, we proposed a relation-contextualized KG embedding model. It represents relations with context-specific embeddings and builds edge translations between entities to preserve KG structures. We proposed context compression and projection to compute edge embeddings. Our experiments on standard datasets demonstrated its effectiveness on entity alignment and link prediction. For future work, we plan to study techniques like language models to represent multi-hop relation contexts. We also want to incorporate other proximity measures into the preserved KG structures, such as attribute similarity.\\

\noindent\textbf{Acknowledgments.} This work is funded by the National Natural Science Foundation of China (No. 61872172), and the Key R\&D Program of Jiangsu Science and Technology Department (No. BE2018131).

\bibliographystyle{splncs04}
\bibliography{reference}

\end{document}